\documentclass[letterpaper, 10 pt, conference]{ieeeconf}
\makeatletter
\let\NAT@parse\undefined
\makeatother

\IEEEoverridecommandlockouts                              %

\usepackage[numbers,sort&compress]{natbib}

\usepackage{multirow}
\usepackage{multicol}
\usepackage[bookmarks=true, hidelinks]{hyperref}
\usepackage{xcolor}
\usepackage{tablefootnote} %
\usepackage{caption}
\usepackage{subcaption}
\usepackage{graphicx}
\usepackage{float}
\usepackage{amssymb}
\usepackage{amsmath}

\usepackage{algorithm}
\usepackage{algpseudocode}
\usepackage{lipsum} 
\usepackage{tabularx}

\usepackage{sidecap}
\usepackage{hyphenat}

\newcommand{\rephrase}[1]{#1}

\usepackage{xspace}

\newcommand{\R}{\mathbb{R}}
\newcommand{\norm}[1]{\left \lVert #1 \right \rVert}

\begin{document}
\bstctlcite{BSTcontrol}

\title{\LARGE \bf
HyperTaxel: Hyper-Resolution for Taxel-Based Tactile Signals \\ Through Contrastive Learning}

\author{Hongyu Li$^{\dagger,\ddagger}$, Snehal Dikhale$^\dagger$, Jinda Cui$^\dagger$, Soshi Iba$^\dagger$, and Nawid Jamali$^\dagger$%
\thanks{$^\dagger$ Honda Research Institute USA. \{snehalsubhash\_dikhale, jinda\_cui, siba, njamali\}@honda-ri.com.
        }%
\thanks{$^\ddagger$ Brown University. This work was done during his internship at Honda Research Institute USA.
        hongyu@brown.edu.}%
}

\maketitle

\begin{abstract}

To achieve dexterity comparable to that of humans, robots must intelligently process tactile sensor data. Taxel-based tactile signals often have low spatial-resolution, with non-standardized representations. In this paper, we propose a novel framework, HyperTaxel, for learning a geometrically-informed representation of taxel-based tactile signals to address challenges associated with their spatial resolution. We use this representation and a contrastive learning objective to encode and map sparse low-resolution taxel signals to high-resolution contact surfaces. To address the uncertainty inherent in these signals, we leverage joint probability distributions across multiple simultaneous contacts to improve taxel hyper-resolution. We evaluate our representation by comparing it with two baselines and present results that suggest our representation outperforms the baselines. Furthermore, we present qualitative results that demonstrate the learned representation captures the geometric features of the contact surface, such as flatness, curvature, and edges, and generalizes across different objects and sensor configurations. Moreover, we present results that suggest our representation improves the performance of various downstream tasks, such as surface classification, 6D in-hand pose estimation, and sim-to-real transfer.

\end{abstract}

\IEEEpeerreviewmaketitle

\section{Introduction}

Tactile sensing is a critical modality for humans to interact with everyday objects~\cite{dahiya_tactile_2010}. Tactile sensors can be divided into two broad categories~\cite{wu_tactile_2022}: vision-based~\cite{yuan_gelsight_2017, donlon_gelslim_2018}, and taxel-based~\cite{tomo_new_2018, wettels_biomimetic_2008, ding_sim--real_2021}. 
Recently, vision-based tactile sensors have gained popularity, partly due to their pixel-based representation, which makes them amenable to deep learning approaches~\cite{li_visualtactile_2023, luu_simulation_2023, yang_touch_2022}. 
However, their size limits full coverage on multi-fingered hands~\cite{li_tata_2022, piacenza_data-driven_2018}. In contrast, taxel-based sensors remain underexplored because they present many challenges to deep learning approaches, including low spatial resolution and a lack of consensus on how to represent and process taxel-based sensors. However, they continue to remain of interest to the robotic manipulation community due to their unique ability to directly respond to the underlying phenomena measured, thereby offering valuable opportunities for enhancing robotic manipulation~\cite{buscher_augmenting_2015, guzey_dexterity_2023}. 

The encoding and processing of taxel signals is still an open research question. Taxel-based sensors present unique challenges before they can be used in downstream tasks. These challenges include 1) the development of effective representations for tactile sensor data, 
and 2) their inherently low resolution~\cite{dahiya_directions_2013, piacenza_data-driven_2018, wu_tactile_2022}, which has been a long-standing barrier hindering tactile dexterous manipulation \cite{dahiya_directions_2013} and perception, such as in-hand 6D pose estimation \cite{li_vihope_2023, dikhale_visuotactile_2022, rezazadeh_hierarchical_2023}. As suggested by \citet{dahiya_directions_2013}, one promising line of research is the use of super-resolution algorithms.

\begin{figure}[t!]
    \centering
    \begin{subfigure}{0.36\columnwidth}
         \includegraphics[width=\textwidth]{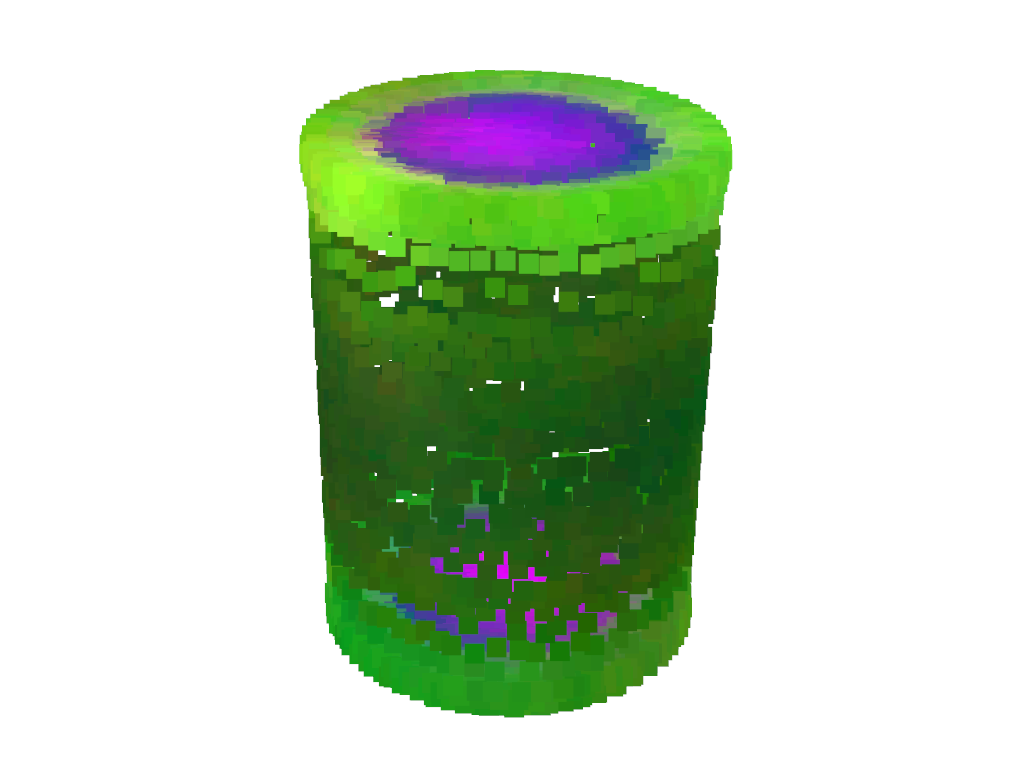}
         \caption{Master Chef Can}
    \end{subfigure}
    \begin{subfigure}{0.36\columnwidth}
         \includegraphics[width=\textwidth]{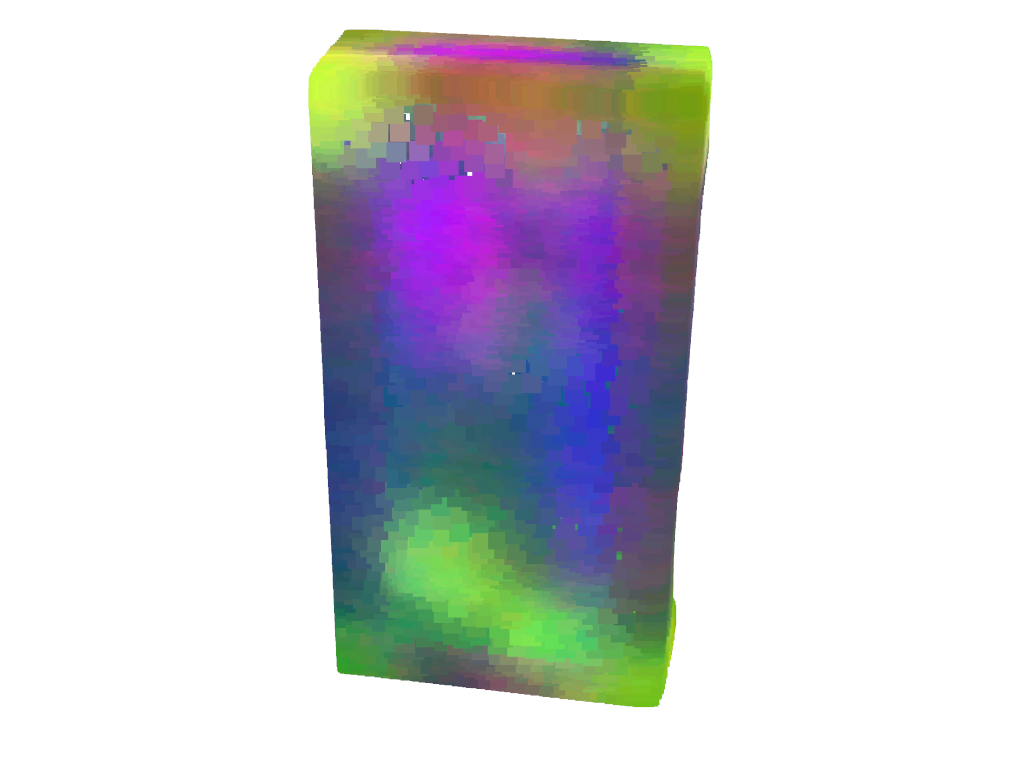}
         \caption{Sugar Box}
    \end{subfigure}
    \begin{subfigure}{0.36\columnwidth}
         \includegraphics[width=\textwidth]{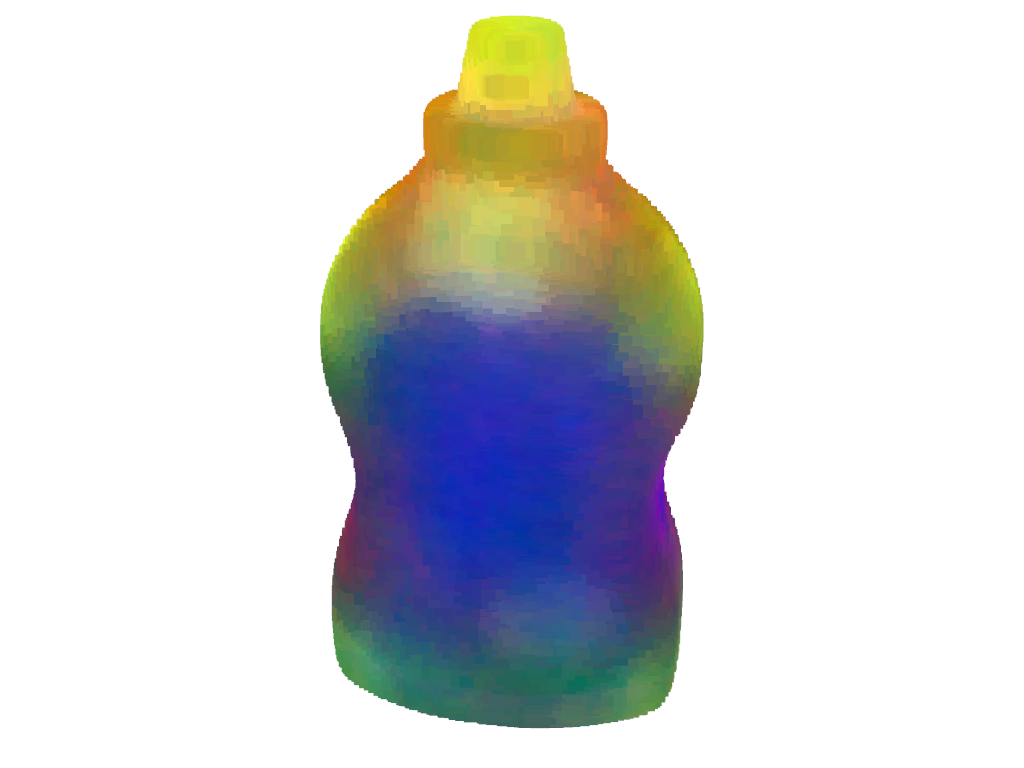}
         \caption{Mustard Bottle}
    \end{subfigure}
    \begin{subfigure}{0.36\columnwidth}
         \includegraphics[width=\textwidth]{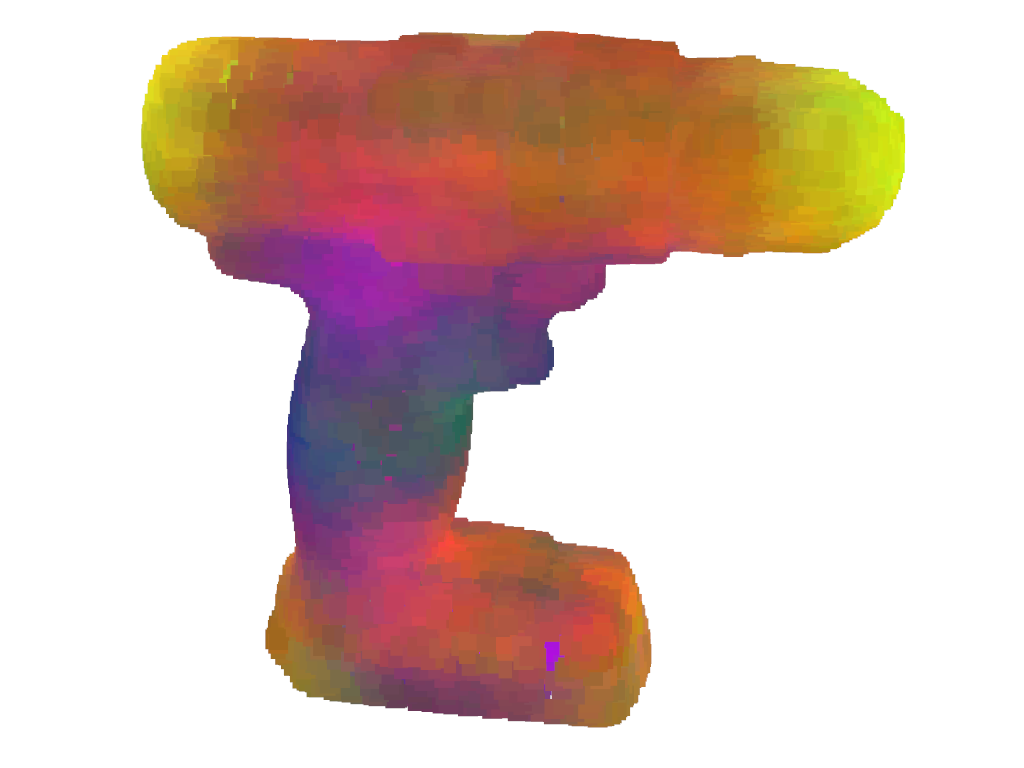}
         \caption{Power Drill}
    \end{subfigure}
    \caption{A visualization of tactile embeddings for different YCB~\cite{calli_ycb_2015} objects. The embeddings capture the geometric features of the contact surface, such as flatness, curvature, and edges, and are consistent across different objects.}
    \label{fig: tac-emb-rgb}
\end{figure}

In this paper, we present a two-stage solution to address the aforementioned challenges.
In the first stage, we propose a method for learning a representation of the tactile signals in an embedding space using contrastive learning. This approach generalizes across various taxel layouts, different objects, and multiple tasks.
Our key intuition is that by exploiting the correspondence between the taxel signals and their contact surface, we can learn a geometrically-informative representation. To this end, we propose graphs to represent the tactile signals, with a novel graph construction strategy and convolution kernel for tactile processing.

In the second stage, we map low-resolution taxel signals to a high-resolution three-dimensional surface using a multi-contact strategy to reduce uncertainty in taxel signals, a process we term hyper-resolution. Unlike super-resolution, which focuses on upscaling data within the same domain, hyper-resolution extends beyond mere upscaling within the same domain across different domains and modalities. For example, in image processing, super-resolution produces the same image with a higher resolution. However, hyper-resolution maps low-resolution taxel signals to capture various object properties, such as the three-dimensional surface of the object, surface texture, etc. This distinction allows hyper-resolution to provide informative data beneficial for tasks such as 6D pose estimation.

Our contributions can be summarized as follows: 1) We propose a novel taxel encoder. 2) We introduce a novel representation learning approach for signals from taxel-based tactile sensors. 3) We propose a novel hyper-resolution algorithm for taxel-based tactile sensors, which leverages the proposed taxel encoder.

 We present results of qualitative analysis that suggest our tactile representation captures geometric features of the contact surface, such as flatness, curvature, and edges, and generalizes across different objects and sensor configurations. Furthermore, we benchmark our proposed framework against two seminal taxel encoders and present quantitative results that demonstrate the effectiveness of our approach. We also perform a comparative analysis across different representations and confirm the effectiveness of the graph representation. We also assess the quality of our hyper-resolution using 6D object pose estimation and show that using our hyper-resolved data improves performance.
 In the end, we verify the sim-to-real transferability of our learned tactile representation on the surface classification task on the real robot.

\section{Related Work}

Representation learning is the process of encoding informative features from raw data to make it suitable for machine learning tasks.
Most of the earlier works use supervised learning methods~\cite{he_deep_2016}.
Recently, there has been a growing interest in self-supervised learning~\cite{grill_bootstrap_2020, he_momentum_2020, he_masked_2021} and multi-modal learning~\cite{radford_learning_2021}.
While most of the prior studies focus on domains such as vision~\cite{grill_bootstrap_2020, he_momentum_2020, he_masked_2021} and language~\cite{radford_learning_2021, brown_language_2020, yang_binding_2024}, in this paper, we are interested in whether the same paradigm can be applied in the tactile domain.

Several prior works have investigated tactile representation learning.
In the image-based sensor domain, \citet{villalonga_tactile_2021} leverage the contrastive framework MoCo~\cite{he_momentum_2020}, and \citet{caddeo_collision-aware_2023} utilize an autoencoder to learn a representation.
However, their transfer from image-based to taxel-based data is non-trivial.
\citet{guzey_dexterity_2023} learn a representation for taxel-based sensors using BYOL~\cite{grill_bootstrap_2020}.
However, their primary focus is on dexterous manipulation, and they do not explore various downstream tasks or representation learning approaches. Therefore, the most effective paradigm for taxel-based signals remains a topic for further exploration.

Contact localization has been utilized to achieve tactile super-resolution. In contact localization, the goal is to estimate the contact location from a given tactile observation. Early works use probabilistic approaches to estimate the probability distribution of the contact location~\cite{corcoran_measurement_2010, luo_localizing_2015, bimbo_-hand_2016, koval_manifold_2017}.
\citet{piacenza_data-driven_2018} adopt a data-driven approach for contact localization in 3D space. 
With the advancements in computer vision, recent works have increasingly explored vision-based tactile sensors~\cite{luo_localizing_2015, villalonga_tactile_2021, caddeo_collision-aware_2023, gao_objectfolder_2023, suresh_midastouch_2023, yang_generating_2023}, which lend themselves well to deep learning techniques due to their pixel-based output. However, the fusion of taxel-based sensors with deep learning methodologies remains relatively underexplored.

In the seminal work~\citet{lepora_tactile_2015} propose taxel-based tactile super-resolution using the Bayesian perception method. 
Recently, there has been a shift from probabilistic-based approaches to learning-based ones.
\citet{wu_tactile_2022} propose a method wherein the taxel-based tactile signal is interpreted as a 2D image, subsequently enhanced using SRGAN~\cite{ledig_photo-realistic_2017}. This process results in a higher-resolution representation of the contact surface between the sensor and the object. However, interpreting tactile signals as a 2D image limits their application to 2D arrangements; the 3D arrangement found in curved fingers cannot be accurately represented. Moreover, they do not utilize the geometric information of the object in contact. %
In this paper, we present a geometrically-informed hyper-resolution algorithm invariant to sensor arrangements.

\section{Methodology}

\begin{figure}[t!]
    \centering
    \vspace*{0.15cm}
    \includegraphics[width=\columnwidth]{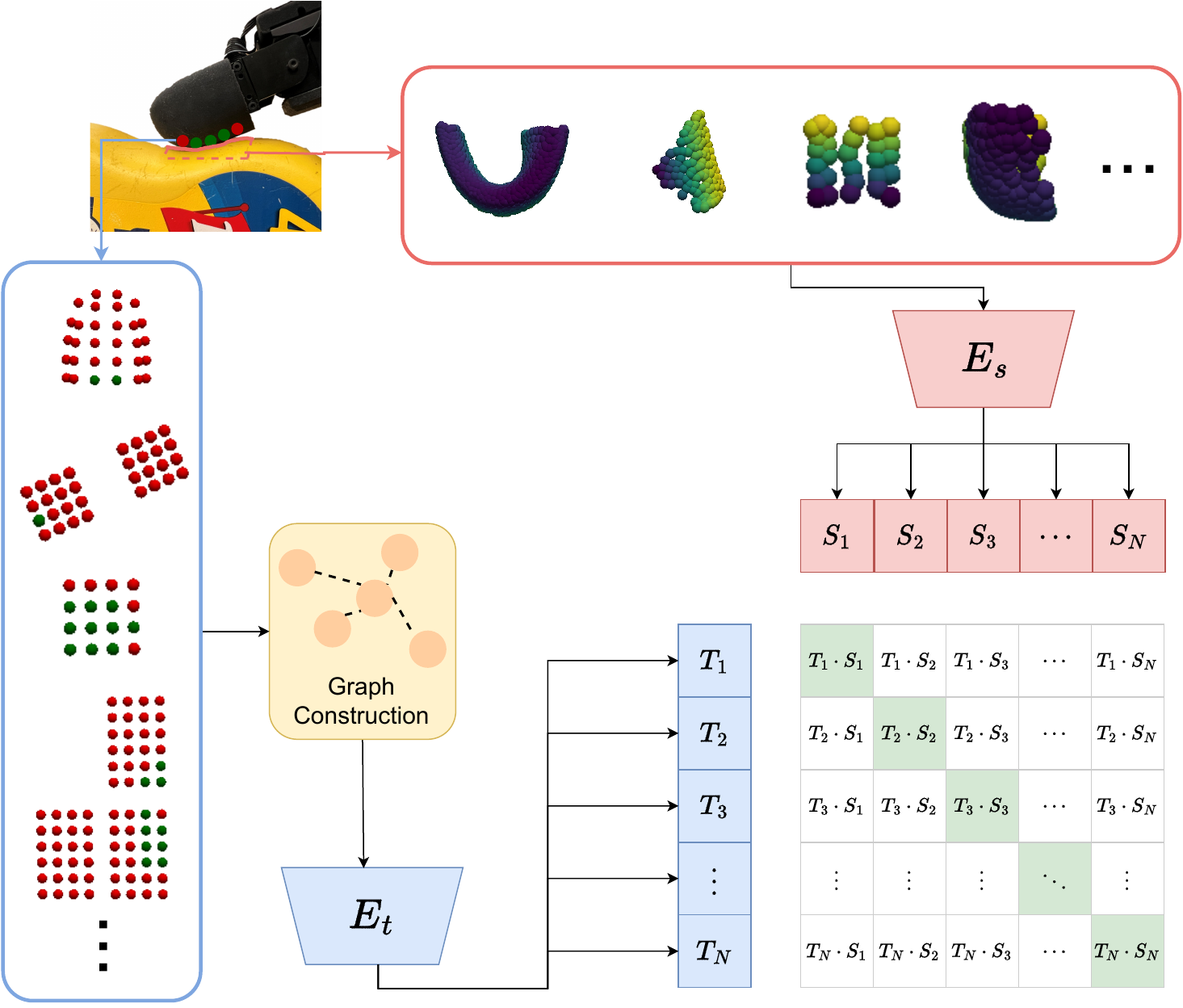}
    \caption{
    Overview of our proposed tactile representation learning framework.
    The tactile signal, left blue box, is represented as a graph and encoded using tactile encoder $E_t$, and the corresponding contact surface patch, top red box, is encoded using surface encoder $E_s$.
    }
    \label{fig:tact-clip}
\end{figure}

Given a sparse taxel-based tactile signal, our goal is to obtain a high-resolution depiction of the contact surface between the sensor and the object of interest. To achieve this, we propose a two-stage solution. The first stage, representation learning, involves using a graph neural network and contrastive learning to learn a geometrically-informed representation of the tactile signals. The second stage, hyper-resolution, uses the learned representation to map low-resolution taxel signals into a high-resolution contact surface using multi-contact localization.

\subsection{Representation Learning}

Figure~\ref{fig:tact-clip} shows an overview of the proposed tactile representation learning framework. In this section, we detail the different components of our framework.

\subsubsection{Taxel Representaion}
\label{sec: taxel_feature}

Tactile data can be represented as point clouds, images, or graphs. The point cloud representation, however, fails to encode the absence of contact, and the image representation struggles to capture the 3D spatial arrangement of tactile sensors. In our research, we opted for the graph representation because it encodes both the spatial arrangement of the taxel signals and the absence of contact. %

We use an undirected spatial graph $\mathcal{G}=(\mathcal{V}, \mathcal{E})$ to represent the taxel data, where $\mathcal{V}$ and $\mathcal{E}$ are the vertices and edges of the graph, respectively. Each vertex corresponds to a taxel of the tactile sensor, and each edge represents the spatial proximity between two taxels. The vertices have two types of features: the 3D coordinates of each taxel, $\mathcal{X} \in \R^{t \times 3}$, and their corresponding signals, $K \in \R^t$, where $t$ is the number of taxels in the tactile sensors. %
For taxel signals with three axes, $K$ is considered the Euclidean norm of the signals from all three axes. We combine $X$ and $K$ into a matrix $\mathcal{V} = [\mathcal{X} | K]$ of dimension $\R^{t \times 4}$. 
To improve sim-to-real transfer, we simplify the taxel signal into a boolean activation state~\cite{yin_rotating_2023, xue_arraybot_2023}.%

Edges $\mathcal{E}$ connect the taxel vertices $\mathcal{V}$ in the taxel graph and construct the graph. Since we are considering tactile sensors with arbitrary spatial arrangements, we propose that tactile message passing should be relative to the spatial distance. We use a radius graph to construct $\mathcal{E}$. In this graph, an edge connects two vertices if their distance is within a certain radius. This approach ensures that the interaction between sensors is stronger when they are closer to each other.

\subsubsection{Tactile Encoder $E_t$}
\label{sec: taxel-encoder}
We process the constructed taxel graph using a graph neural network (GNN).
Specifically, the taxel graph passes through three message passing layers~\cite{gilmer_neural_2017}, a pooling layer, and a non-linear layer output head.
The message-passing layer is defined as
\begin{equation}
    n_i' = \gamma(n_i, \bigoplus_{j \in \mathcal{N}(i)} \phi(n_i, n_j, e_{j,i})) ,
\end{equation}
where $n_i$ is the vertex feature, and $e_{j,i}$ is the edge feature between vertices $i$ and $j$.
$\bigoplus_{j \in \mathcal{N}(i)}$ is a differentiable aggregation function, such as maximum or summation, and $\gamma$ and $\phi$ are two differentiable functions such as MLPs.
Our observation is that the taxel signals rely on relative features with respect to their neighbors instead of absolute features.
For example, a $4\times4$ taxel pad with evenly high activation signals and evenly low activation signals should represent the same contact surface (flat surface).
Therefore, we propose to use the EdgeConv operator~\cite{wang_dynamic_2019}, which leverages the relative features between vertices $n_i$ and $n_j$
\begin{equation}
    n_{i}' = \sum_{j \in \mathcal{N}(i)}(\phi(x_i, h_i, x_j - x_i, h_j - h_i)),
\end{equation}
where $h_i$ and $h_j$ are the hidden features of node $i$ and $j$.

\subsubsection{Sensor-Object Contact Surface Representation}
\label{sec: taxel-hyper-resolution-representation}
To map the low-resolution tactile signals to the high-resolution surface shape, we need to represent the contact surface between the sensor and the object. To this end, we represent the sensor-object contact surface as a cube encapsulating the object's surface that is in contact with the sensor. The cube’s height and width are set to the sensor’s dimensions, and its depth, $\delta_p$, represents the penetration into the object’s surface. In our experiments, we set $\delta_p = 0.8~cm$, as lower values of $\delta_p$ increased the risk of mesh collision during initialization.
The intersection between the object $\mathcal{O}$ and this cube is referred to as a contact surface patch. 
The contact surface patch captures the local geometry of the object at the contact point, and can be used to learn a correspondence between the low-resolution tactile signal and the high-resolution object surface shape. 
We view the contact surface patch as the hyper-resolution space of that respective tactile sensor.

\subsubsection{Learning the Tactile Representation}

We propose the contrastive learning framework shown in Fig. \ref{fig:tact-clip} to learn the tactile representation that leverages the inherent relationship between the contact surface and the tactile signals.
For example, when the sensor is pressed against a flat surface, the taxel signals should demonstrate the flatness feature.
On the contrary, when the sensor is pressed against a curved surface, the signals should demonstrate the curvature feature and distinguish itself from the flatness feature. %
Our key insight for learning an effective representation of tactile signals is to exploit this correspondence.

Inspired by the vision-language learning framework CLIP~\cite{radford_learning_2021}, we draw $N$ random pairs of tactile sensor signals (the blue box) and corresponding contact surfaces (the red box) for each data sample. 
While the CLIP framework is typically used for visual-language tasks, we adapt it for tactile representation learning.
This adaptation requires two significant modifications to the original formulation.
1) Due to the lack of an existing dataset for taxel-based tactile sensors, we need to collect the required paired data.
2) Since the original encoders (vision and language) are incompatible with taxel signals, we need to design a neural network model to encode taxel signals.

The tactile graph described in Section~\ref{sec: taxel_feature} is passed to the tactile encoder $E_t$ (Sec.~\ref{sec: taxel-encoder}) and encoded into tactile embedding $T \in \R^{n}$, where $n$ represents the embedding size. 
Second, the contact surface data (red box), represented as a point cloud, is encoded through the surface encoder $E_s$ into surface embedding $S \in \R^{n}$, which has the same size as the tactile embedding $\R^{n}$. 
We choose PointNet \cite{qi_pointnet_2017} as our surface encoder.
The embedding size $n$ is empirically set as 128.

The final step involves learning tactile representation. This is achieved by computing the dot product of the tactile embedding and surface embedding $T \cdot S^\intercal $,  resulting in a $N \times N$ matrix as depicted in the bottom right corner of Fig. \ref{fig:tact-clip}. 
This matrix contains $N$ positive pairs and $N^2-N$ negative pairs. %
The dot product operation measures the cosine similarity between the tactile embedding and surface embedding. 
We optimize both encoders using a symmetric cross-entropy loss~\cite{radford_learning_2021} such that the $N \times N$ matrix turns into an identity matrix $\mathbb{I}^{N \times N}$. 
By doing so, we learn a representation that brings matching pairs closer together and pushes non-matching pairs farther apart in the embedding space.

\subsection{Multi-Contact Localization for Hyper-Resolution}
\label{sec: multi-contact-local}

\begin{figure}[t!]
    \centering
    \vspace*{0.15cm}
    \includegraphics[width=\columnwidth]{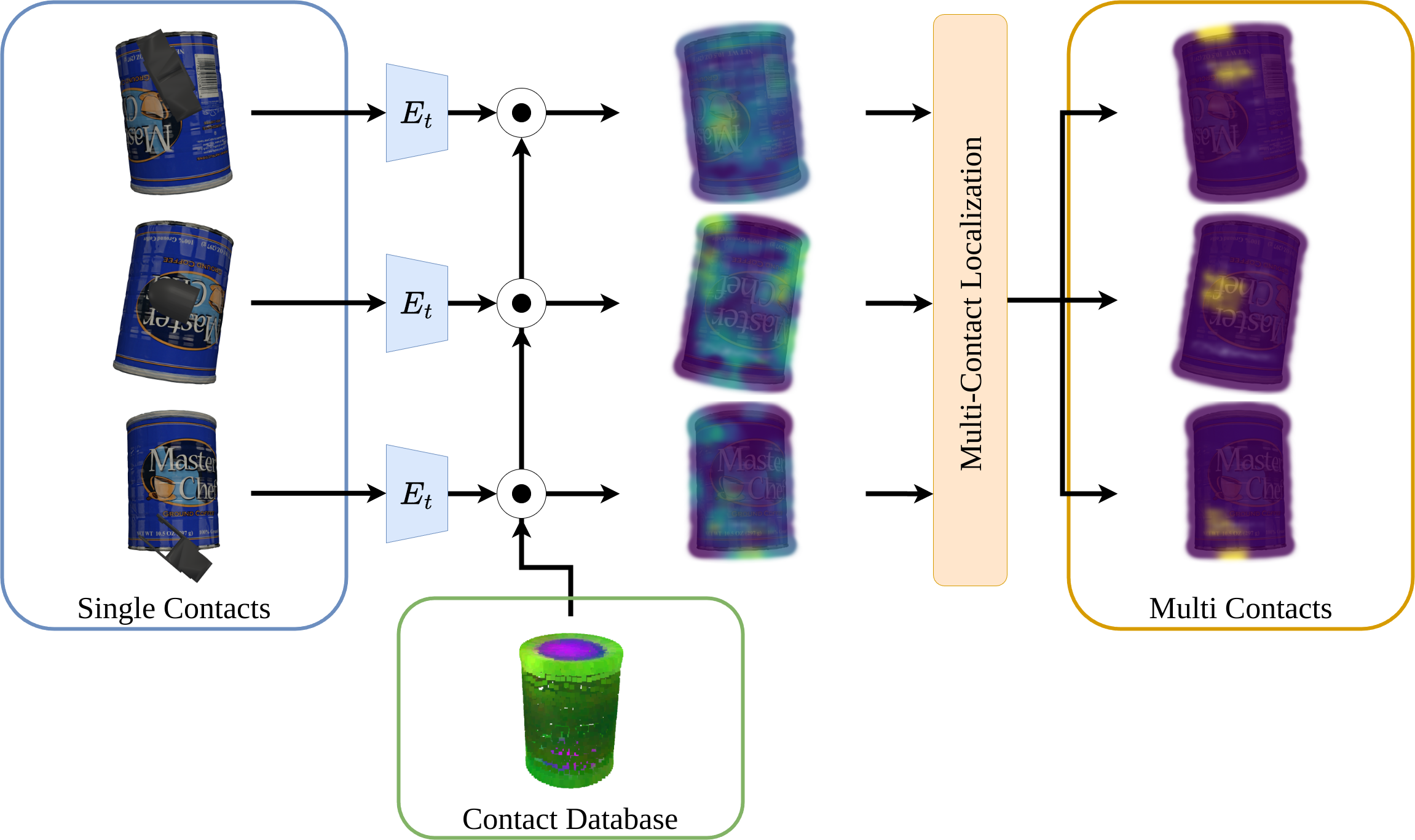}
    \caption{A visual illustration of multi-contact localization. The first column shows the contact locations. The second and third columns show the likelihood map of contact location using single-, and multi-contact reasoning, respectively.}
    \label{fig: multi-contact-diag}
\end{figure}

We develop an innovative approach that used multi-contact localization to transform sparse touch data into detailed object surface geometry, thereby achieving hyper-resolution from tactile sensors with limited spatial resolution. Consider a collection of tactile patterns and their corresponding object surface details, analogous to a library of tactile experiences. Given the spatial sparsity inherent in these tactile patterns, confidently associating a tactile pattern with its object surface is a non-trivial task. To address this challenge, we reason over multiple simultaneous contacts with the object to increase confidence in our estimates. %

Having a set of objects $\mathcal{O}$, we first collect a contact database $\mathcal{B}_o$ for each object $o \in \mathcal{O}$ offline, which consists of the contact surface patches $\mathcal{S}_{o,c}$, the corresponding contact signals $\mathcal{C}_o$, and the 6D poses of the sensor $\mathcal{P}_{o,c}$ in a common frame of reference $F_c$. %
We omit the subscript $o$ in the following paragraph for simplicity and note that these data are object $o$ specific.
For each sample $b_i$ in the database $\mathcal{B}_o$, $b_i = (s_{c,i} \in \mathcal{S}_c, c_i \in \mathcal{C}_o, p_{c,i} \in \mathcal{P}_c)$.
The 6D pose $p_{c,i} \in \R^7$ is represented as the concatenation of 3D translation $\R^3$ and 3D rotation in quaternion form $\R^4$.
To ease online computation, we preprocess the contact surface patches $\mathcal{S}_c$ and encode them into embeddings $\mathcal{S}_l$ using the pretrained surface encoder $E_s$.

During deployment, we assume there is $N_c$ number of sensors that are in contact with the object $o$.
We denote $\mathcal{J}$ as the set of these sensors and the actual pose of each sensor $j \in \mathcal{J}$ as $\mathcal{P}_{u,j} \in \R^7$.
The robot’s forward kinematics is used to transform the sensor poses to a common frame of reference.
Each sensor $j$ has a sensor reading $d_j$ represented as the taxel graph (Sec.~\ref{sec: taxel_feature}).
We encode the collection of sensor readings $\mathcal{D} = \{d_i \mid i \in 1,2, \cdots, N_c\}$ into taxel embeddings $\mathcal{T}$ using tactile encoder $E_t$ such that $\mathcal{T} = \{ T_i = E_t(d_i) \mid \forall d_i \in \mathcal{D} \}$.
We measure the similarity between $\mathcal{T}$ and the surface embeddings $\mathcal{S}_l$ stored in the database and rank the candidate poses $\mathcal{P}_c$ accordingly.
We take top $\delta_C$ candidates to reduce the computation in later steps and obtain a distribution of contact locations $\Omega_i \subseteq \mathcal{P}_c$~\cite{caddeo_collision-aware_2023}. 

For any two sensors in contact $j_a, j_b \in \mathcal{J}$, we obtain their respective distribution $\Omega_a, \Omega_b$ for contact location candidates. 
We filter the pair-wise Euclidean distance between each candidate location $p_{c,a} \in \Omega_a, p_{c,b} \in \Omega_b$ using the actual sensor poses $\mathcal{P}_a, \mathcal{P}_b$ 
\begin{equation}
\label{eqn: dist}
\begin{aligned}
    \Omega_{d,a} &= \{p_{c,a} \mid \norm{p_{c,a} - p_{c,b}} - \norm{\mathcal{P}_a - \mathcal{P}_b} \leq \delta_n \} \\
    \Omega_{d,b} &= \{p_{c,b} \mid \norm{p_{c,a} - p_{c,b}} - \norm{\mathcal{P}_a - \mathcal{P}_b} \leq \delta_n \} ,
\end{aligned}
\end{equation}
resulting in two distance filtered sets $\Omega_{d,a}$ and $\Omega_{d,b}$.
$\delta_n$ is a threshold to offset noises, such as in calibration and forward kinematics.
Repeating this operation on all $N_c$ sensors in contact, we obtain $N_c$ distance filter sets $\Omega_{d,1}, \Omega_{d,2}, \cdots, \Omega_{d,N_c}$.

We desire to find an optimal solution $\Psi \in \Omega_{d,1} \times \Omega_{d,2} \times \cdots \times \Omega_{d,N_c}$ that maximizes the similarities between taxel embeddings $\mathcal{T}$ and the surface embeddings $\mathcal{S}_l$.
To achieve this, we build a multipartite graph $K$ with $N_c$ partites. 
Each candidate left in $\Omega_d$ is added as a node, and edges are added if the distance constraint (Eqn.~\ref{eqn: dist}) is satisfied. 
We use Paton's algorithm~\cite{paton_algorithm_1969} to find the cycle $\Psi$ with the largest joint probability. 

Figure~\ref{fig: multi-contact-diag} illustrates the process of Hyper-Resolution.

\section{Datasets}
We collected two datasets using NVIDIA Isaac Sim for a subset of YCB objects~\cite{calli_ycb_2015}. %
We used an Allegro Hand equipped with XELA tactile sensors, resembling our real-world setup. The Allegro Hand has eleven $4\times4$ flat pad sensors, three $4\times6$ flat pad sensors, and four curved tip sensors (each has 30 taxels). 
The first dataset, Section~\ref{sec: contact-db}, is a comprehensive database of tactile sensors interacting with the objects. This dataset serves as our tactile experience library and is used to evaluate tactile representation learning.
The second dataset, Section~\ref{sec: 6d-pose-dataset}, consists of the Allegro Hand holding an object and executing random trajectories. This dataset is used to evaluate the performance of our methods on a downstream task, namely the in-hand 6D pose estimation task.

\subsection{Contact Database}
\label{sec: contact-db}
\begin{figure}[t!]
\vspace*{0.15cm}
    \centering
    \includegraphics[width=\columnwidth]{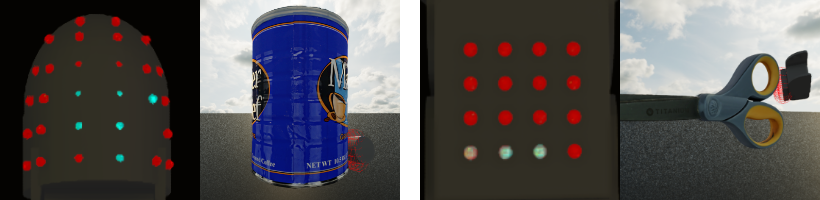}
    \caption{Illustration of data collection using NVIDIA Isaac Sim simulator. The figure shows the curved fingertip sensor (left) and the $4\times4$ flat pad sensor (right), respectively.}
    \label{fig:contact-db}
\end{figure}

We first construct a dataset that captures tactile experiences across the entire surface of an object at various points. The tactile sensor is simulated using the \texttt{Contact Sensor} provided by Isaac Sim.
We sample 2048 points on the object mesh using Poisson disk. 
Each point corresponds to a 3D position $\R^3$ and its respective surface normal $\R^3$.
Like previous works~\cite{villalonga_tactile_2021, suresh_midastouch_2023, caddeo_collision-aware_2023}, we randomly chose a subset of points to collect our tactile experience.
For each selected point, we align the tactile sensor's z-axis with the surface normal and conduct eight contact trials. 
In each trial, the sensor is rotated 45$^{\circ}$. We start by positioning the sensor 2.5 cm away from the point and then gradually push it towards the surface along the normal. 
Once the sensor is in contact with the object, we collect the tactile observations and their corresponding poses.
This process is repeated for each type of taxel sensor on the Allegro Hand, which includes $4\times4, 4\times6$, and curved tips, to compile a comprehensive contact database.

\begin{figure}[t!]
    \centering
    \vspace*{0.15cm}
    \begin{subfigure}{0.48\columnwidth}
         \centering
         \includegraphics[width=\textwidth]{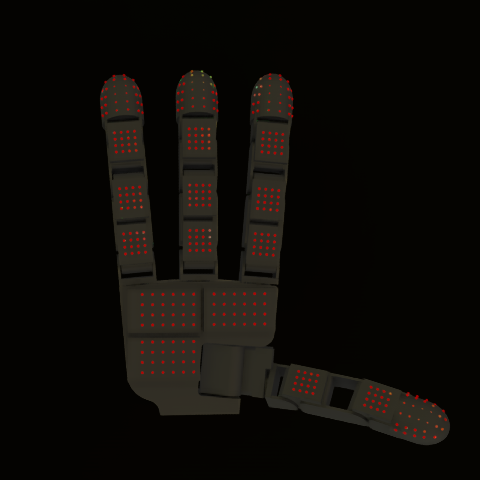}
         \caption{}
         \label{fig: allegro-taxel}
    \end{subfigure}
    \begin{subfigure}{0.48\columnwidth}
         \centering
         \includegraphics[width=\textwidth]{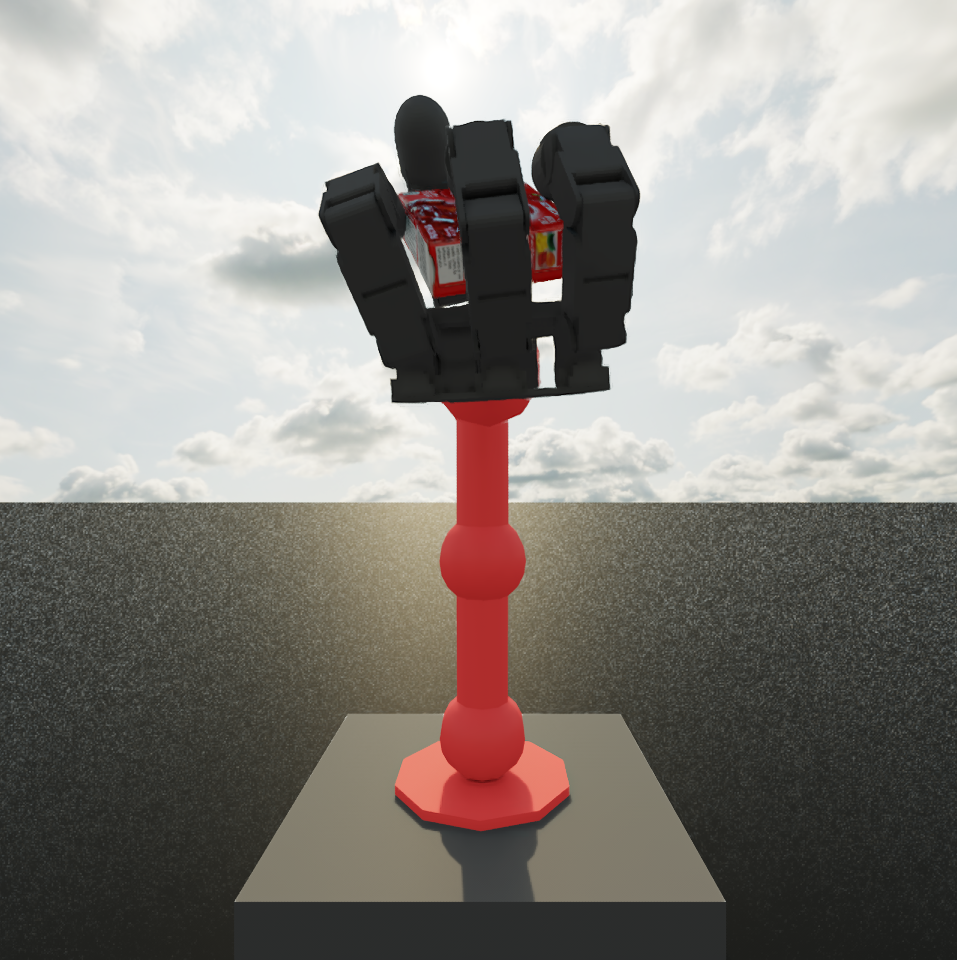}
         \caption{}
         \label{fig: allegro-grasp}
    \end{subfigure}
    \caption{Allegro Hand equipped with XELA tactile sensors: a) taxel distribution, b) grasping an object  }
\end{figure}

\subsection{In-hand Object Dataset}
\label{sec: 6d-pose-dataset}

To evaluate our framework on downstream tasks such as 6D in-hand pose estimation, we also collected a simulated dataset, which consists of the Allegro Hand holding an object and executing random trajectories. 
For each object, we collect 16,000 samples for the training set and 4,000 samples for the validation set.
The dataset is collected using the following procedures:  
\begin{enumerate}
    \item The hand is initialized to face upwards.
    \item The object is dropped from a height of 2$cm$ above the hand, with its pose randomly initialized.
    \item The hand performs the grasping action.
    \item If the grasp fails (e.g., the object falls), return to step~1.
\end{enumerate}

\section{Experiments}

We utilize the AdamW~\cite{loshchilov_decoupled_2018} optimizer with a learning rate of 0.001. We pre-train the tactile encoder for 100 epochs using all objects and optimize the pose estimation model for each object for 500 epochs.

We begin by qualitatively assessing the tactile representation learned through our approach (Section~\ref{sec: qualitative-emb}). Next, we examine the effectiveness of our method in the hyper-resolution task (Section \ref{sec: hyper-resolution}). We then ablate the chosen graph operators and constructors (Section~\ref{sec: ablation}) and study the impact of multi-contact localization on hyper-resolution (Section \ref{sec: multi-contact-loc-hyper-res}).  In addition, we test our approach on two downstream tasks: in-hand object pose estimation (Section~\ref{sec: 6d-pose}) and surface classification (Section \ref{sec: surface-classification}).%

\subsection{Qualitative Analysis of Learned Tactile Representation}
\label{sec: qualitative-emb}

To evaluate the quality of our learned tactile representation, we performed a qualitative analysis using visualizations of the tactile embeddings. We used the contact database (Section \ref{sec: contact-db}) to generate the tactile embeddings for each contact using our learned tactile encoder. We then applied principal component analysis (PCA) to reduce the dimensionality of the embeddings to 3, and used the resulting values as RGB colors for visualization.

The results, as depicted in Fig.~\ref{fig: tac-emb-rgb}, reveal that the tactile embeddings effectively capture the geometric features of the contact surface, such as flatness, curvature, and edges. For example, the flat surfaces on the Master Chef Can, Sugar Box, and Mustard Bottle are all represented in shades of purple and blue, while the curved surfaces are depicted in yellow and green. The edges of the Master Chef Can are highlighted in lime green, indicating a stark contrast between the neighboring points. Notably, the tactile embeddings demonstrate consistency across different objects and sensor types, demonstrating the generalization of our representation.

\subsection{Hyper-Resolution Performance Evaluation}
\label{sec: hyper-resolution}

\begin{table}[t!]
\caption{Comparison of Hyper-Resolution Performance using Different Tactile Representations.}
\label{tab: taxel-encoder}
\centering
\begin{tabular}{c | r | r} 
 \hline
 Method & Rank $\downarrow$ & CD $\downarrow$ \\
 \hline
 Image-based (CNN) & 104.40 & 0.82 \\
 Point cloud-based (PointNet) & 150.31 & 0.85 \\
 \textbf{Graph-based (Ours)} & \textbf{84.05} & \textbf{0.74} \\
 \hline
\end{tabular}
\end{table}

In this section, we evaluate the performance of our proposed hyper-resolution algorithm, which maps the low-resolution taxel signals to high-resolution contact surface patches using a contact database. We compare our method with two baselines: image-based approaches (CNN) \cite{wu_tactile_2022, guzey_dexterity_2023}, and point-cloud-based approaches (PointNet) ~\cite{dikhale_visuotactile_2022, rezazadeh_hierarchical_2023}. 

We use two metrics to measure the accuracy of our hyper-resolution: Chamfer distance (CD) and rank. Chamfer distance computes the average minimum distance between two point sets, and reflects the geometric similarity between the estimated and ground truth contact surfaces. Rank measures the precision of identifying the correct surface based on the similarity between the tactile embeddings and the surface embeddings. The rank of an algorithm is then the average rank it assigns to the ground truth surface across all tactile contact points in the database in Section \ref{sec: contact-db}. A lower rank means a better performance. Table~\ref{tab: taxel-encoder} shows the results of these experiments. We observe that our method outperforms both of the baselines on both metrics, demonstrating the effectiveness of our hyper-resolution algorithm.

\subsection{Comparison of Graph Operators and Constructors}
\label{sec: ablation}

\begin{table}[t!]
\caption{Performance of Different Graph Operators and Constructors on the Hyper-Resolution Task.}
\label{table:graph-kernels}
\centering
\begin{tabular}{c | c | r | r} 
 \hline
 Graph Operator & Graph Constructor & Rank $\downarrow$ & CD $\downarrow$ \\
 \hline
 TacGNN & KNN ($n=1$) & 111.03 & 0.99 \\
 TacGNN~\cite{yang_tacgnn_2023}  & KNN ($n=3$) & 87.11 & 0.79 \\
 TacGNN & KNN ($n=5$) & 114.10 & 1.09 \\
 TacGNN & Radius ($r=0.005$) & 201.06 & 2.47 \\
 TacGNN & Radius ($r=0.01$) & 115.71 & 1.12 \\
 TacGNN & Radius ($r=0.015$) & 117.57 & 1.10 \\
 GCN~\cite{garcia-garcia_tactilegcn_2019}  & KNN ($n=3$) & 111.77 & 0.89 \\
 EdgeConv & KNN ($n=1$) & 92.31 & 0.81 \\
 EdgeConv & KNN ($n=3$) & 84.44 & 0.75 \\
 EdgeConv & KNN ($n=5$) & 84.85 & 0.75 \\
 EdgeConv & Radius ($r=0.005$) & 168.37 & 2.25 \\
 \textbf{EdgeConv (our)} & \textbf{Radius ($r=0.01$)} & \textbf{84.05} & \textbf{0.74} \\
 EdgeConv & Radius ($r=0.015$) & 85.21 & 0.76 \\
 \hline
\end{tabular}
\end{table}
We ablate the impact of different graph operators and constructors on the quality of the learned tactile representation. We compared our proposed EdgeConv operator with two seminal works: TacGNN~\cite{yang_tacgnn_2023}, GCN~\cite{garcia-garcia_tactilegcn_2019}. We also examined different graph constructors, such as KNN and radius graphs, with different parameters. Table~\ref{table:graph-kernels} shows the results of these experiments. We observe that our EdgeConv operator combined with the radius graph constructor (r = 0.01) achieves the best performance in terms of rank and CD metrics. This indicates that our operator can effectively capture the relative features between taxels and that the radius graph can better reflect the spatial distance between~taxels.

\subsection{Effect of Multi-Contact Localization on Hyper-Resolution}
\label{sec: multi-contact-loc-hyper-res}

In this section, we evaluate how the number of contacts affects our hyper-resolution algorithm. Taxel-based sensors perceive coarser geometry features, making it challenging to estimate the corresponding surface from a single observation. Previous studies \cite{villalonga_tactile_2021, caddeo_collision-aware_2023} have confirmed performance gains by incorporating multi-contacts on vision-based tactile sensors. In this study, we extend this concept to taxel-based sensors. 

\begin{figure}[t!]
    \centering
    \includegraphics[width=0.6\columnwidth]{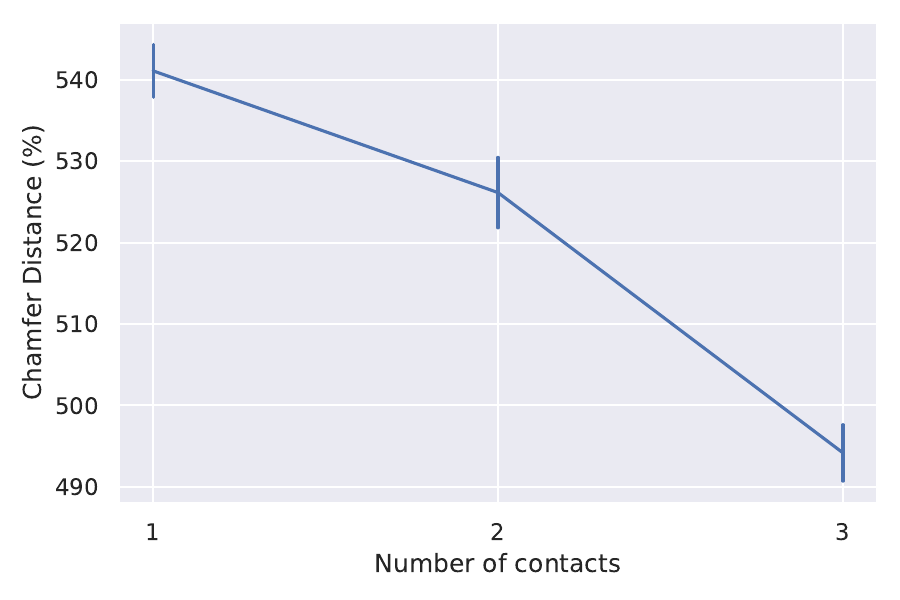}
    \caption{Effect of Multi-Contact Localization on Hyper-Resolution.}
    \label{fig: multi-contact-numbers}
\end{figure}

Figure~\ref{fig: multi-contact-numbers} shows the results of the quantitative analysis of this experiment. We observe that the CD decreases as the number of contacts increases, indicating that the hyper-resolution quality improves with more contacts. This is because more contacts provide more information and constraints about the object surface, reducing the ambiguity and uncertainty in the hyper-resolution. 

\rephrase{
We provide a visualization sample in Fig.~\ref{fig: multi-contact-diag}. 
The second and third column shows the likelihood map of a single contact and multi-contact with the object in each row, respectively. 
Brighter colors indicate a higher likelihood. We notice the curved surfaces are accurately depicted with brighter colors, suggesting that the algorithm correctly identifies these contacts as originating from a curved surface. The third column shows the refined likelihood map after applying multi-contact reasoning. After multi-contact reasoning, the true contact areas are brightly colored while all other areas are dark, accurately pinpointing the potential origin of the tactile input on the object.
}

\subsection{Effect of HyperTaxel on In-Hand 6D Pose Estimation}
\label{sec: 6d-pose}

\begin{figure}[t!]
\vspace*{0.15cm}
\centering
    \includegraphics[width=\columnwidth]{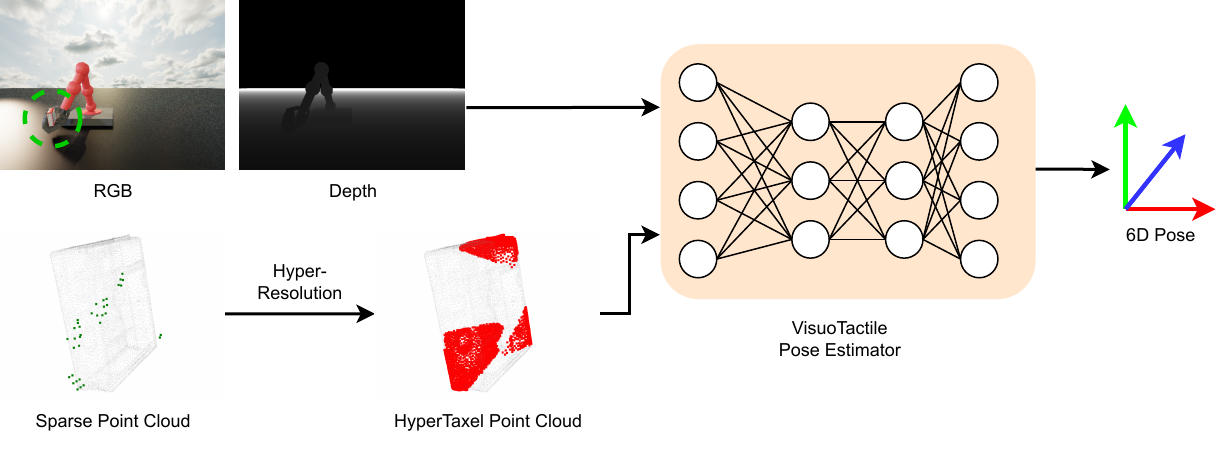}
    \caption{An illustration of the modified ViTa algorithm with our hyper-resolution module. The tactile data is enhanced by our module to produce a high-resolution representation of the object surface. This representation is then fed into the ViTa algorithm as its tactile input.}
    \label{fig: pose-estimator-pipeline}
\end{figure}

In this section, we evaluate the effectiveness of our approach by integrating it with ViTa~\cite{dikhale_visuotactile_2022}, an existing visuotactile model for 6D pose estimation. ViTa uses visual and tactile data to represent the object’s surface, but low-resolution tactile data can affect its performance. Fig.~\ref{fig: pose-estimator-pipeline} shows the modified pipeline, which includes our hyper-resolution method to map the sparse low-resolution tactile data to a high-resolution object surface representation. This representation is then fed into the ViTa algorithm without any further changes.
Following prior works~\cite{dikhale_visuotactile_2022, li_vihope_2023}, we evaluate the performance using three metrics: position error (cm), angular error (deg), and ADD (cm). Position error is the L2 norm of the difference between the estimated and ground truth translation vectors, $||t-\hat{t}||_2$. Angular error is the inverse cosine of the inner product of the estimated and ground truth quaternions, $\cos^{-1}(2 \langle R, \hat{R} \rangle^2 -1)$, and ADD measures the pairwise distances between the 3D model points transformed using estimated and ground truth 6D poses, $\frac{1}{m} \sum_{x \in o} ||(Rx+T)-(\hat{R}x+\hat{T})||$, where $x$ is the 3D point, and $m$ is the number of 3D points on the object model $o$.

\begin{table}[t!]
\caption{Comparative Analysis of the Effect of HyperTaxel on In-Hand 6D Pose Estimation.}
\label{tab: pos-est-table}
\centering
\begin{tabular}{lrrr}
\hline
Method          & Angular Error & Position Error & ADD \\
\hline
DenseFusion   & 10.52 $\pm$ 0.12 & 0.46 $\pm$ 0.00 & 0.87 $\pm$ 0.01 \\
ViTa          &  8.85 $\pm$ 0.10 & 0.43 $\pm$ 0.00 & 0.77 $\pm$ 0.01 \\
\textbf{ViTa+HyperTaxel} &  \textbf{8.56 $\pm$ 0.10} & \textbf{0.40 $\pm$ 0.00} & \textbf{0.74 $\pm$ 0.01} \\
\hline
\end{tabular}
\end{table}

\begin{figure}[t!]
    \centering
    \vspace*{0.15cm}
    \begin{subfigure}{.9\columnwidth}
        \includegraphics[width=\textwidth, height=5cm]{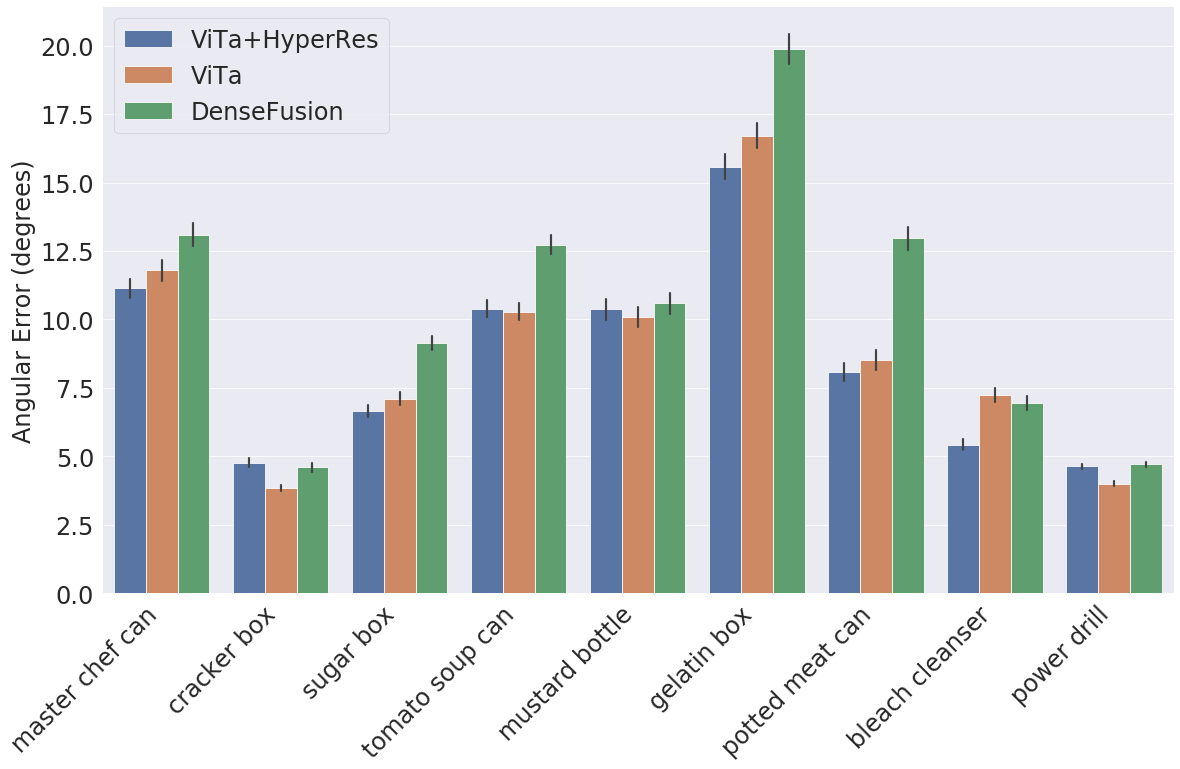}
    \end{subfigure}
    \begin{subfigure}{.9\columnwidth}
        \includegraphics[width=\textwidth, height=5cm]{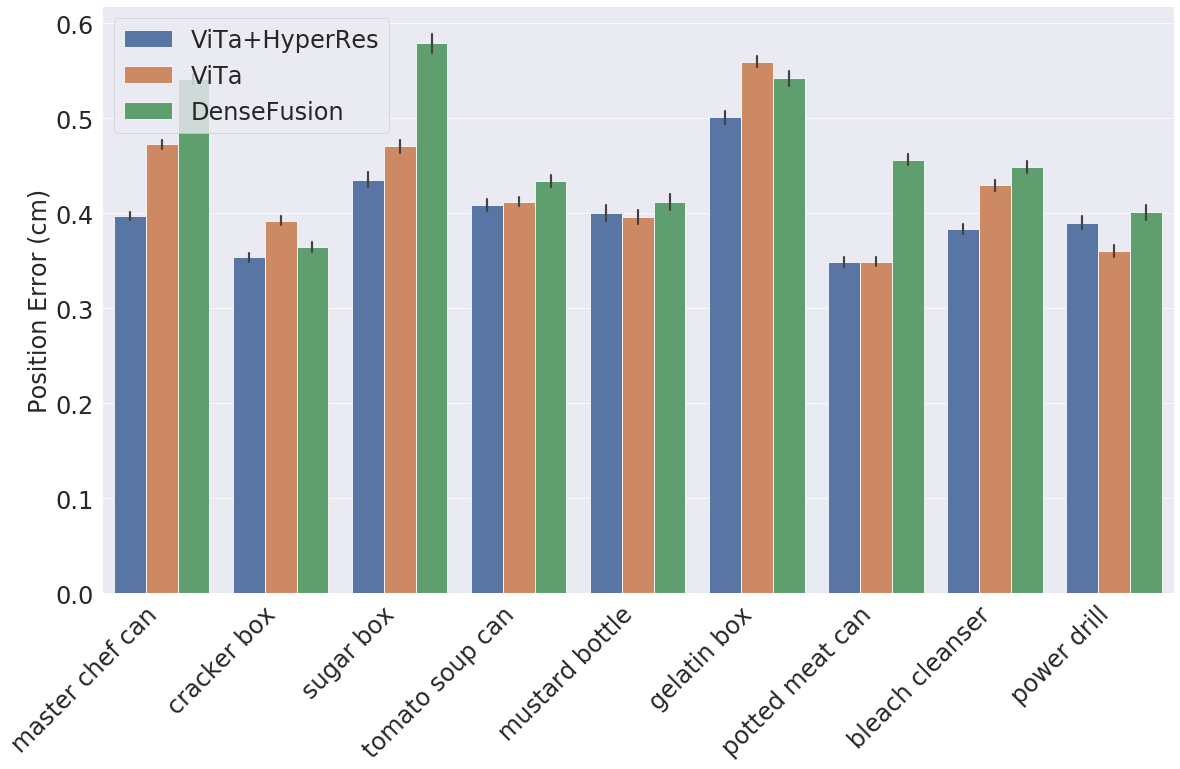}
    \end{subfigure}
    \caption{Pose estimation performance on YCB objects. }
    \label{fig: pos-est-bar}
\end{figure}

We verify the performance of our proposed hyper-resolution algorithm in the synthetic pose estimation data collected in Sec.~\ref{sec: 6d-pose-dataset}. Table~\ref{tab: pos-est-table} shows the results.
We first compare the vision-only baseline DenseFusion~\cite{wang_densefusion_2019} with the visuotactile baseline ViTa~\cite{dikhale_visuotactile_2022}.
By adding tactile information, ViTa has a 1.67 degrees lower angular error and 0.03~$cm$ lower position error. ViTa+HyperTaxel outperforms all of them. 
\rephrase{An object-wise analysis, depicted in Fig.~\ref{fig: pos-est-bar}, reveals enhancements for most objects.
Notably, our approach faces challenges on power drill objects which reveals one potential limitation.
Since our offline collected database relies on random sampling on the object model, our current choice of sample number might not capture the complex geometry of the power drill accurately.
In the future, this limitation might be lifted by scaling up the samples on the object.
}

\subsection{Real Robot Results}

We deployed our model, trained on synthetic data on a multi-fingered gripper (Allegro Hand equipped with XELA tactile sensors) affixed to a Sawyer robot. The tactile sensors capture surface contact points on the object. 
We use real YCB objects to evaluate the performance of our framework in a real-world robot environment.

\label{sec: surface-classification}

\begin{figure}[t!]
    \centering
    \begin{subfigure}{0.3\columnwidth}
         \centering
         \includegraphics[width=\textwidth]{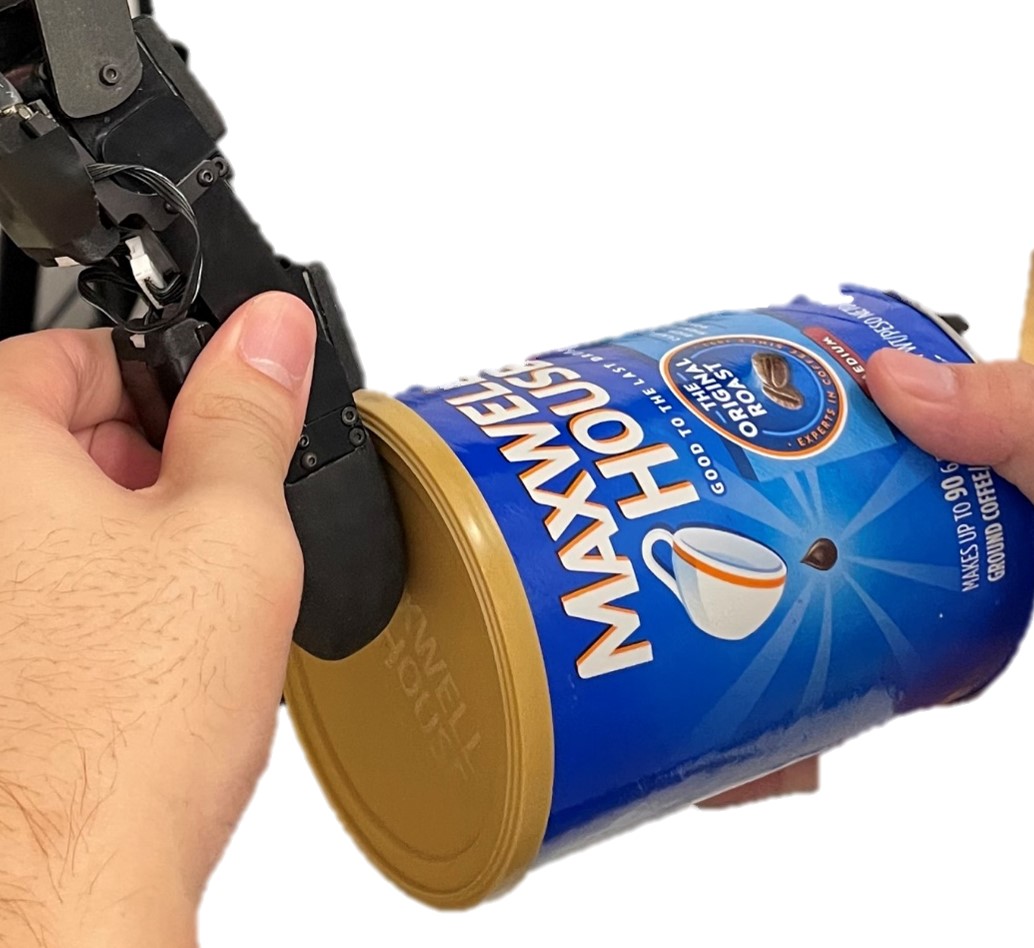}
         \label{fig: surface-flat}
    \end{subfigure}
    \begin{subfigure}{0.3\columnwidth}
         \centering
         \includegraphics[width=\textwidth]{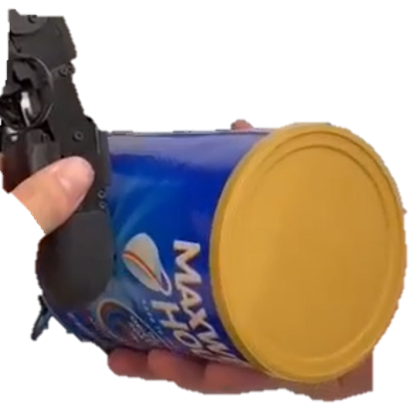}
         \label{fig: surface-curve}
    \end{subfigure}
    \caption{Demonstration of data collection process. We collect a real-world dataset by pressing the tactile sensor-equipped fingertip on flat surfaces (left) and curved surfaces (right).}
    \label{fig: surface-collection}
\end{figure}

A good representation should demonstrate ability to distinguish different surface types. We evaluate our representation on the surface classification task to demonstrate its ability to cluster tactile signals based on the geometric features of the contact surfaces, such as flatness and curvature. To conduct this experiment, we use a real-world object that has both flat and curved surfaces: the Master Chef can. As shown in Fig. \ref{fig: surface-collection}, we press the tactile sensors (both the curved tip and the 4×4 flat pad) on different parts of the can. We collect the tactile signals from multiple contacts on each surface, covering the flat and curved areas as evenly as possible. \rephrase{A video demonstration of this process is available in our supplementary material.}

We then encode the tactile signals using three representations: BYOL, AE, and ours. We also compare against directly using the raw data (Raw). We then apply the K-means clustering algorithm to classify them into two classes: flat and curved. Table~\ref{tab: surface_class} shows the results of this experiment, where we measure the performance of our representation using three metrics: random index (RI), adjusted random index (ARI)~\cite{hubert_comparing_1985}, and accuracy. RI measures the similarity between the estimated clustering and the ground-truth clustering. ARI takes into account the expected value of the RI, which is the random guessing probability. 
We include the accuracy metric, following the linear classification protocol used to evaluate representation quality in self-supervised learning~\cite{grill_bootstrap_2020, he_masked_2021, he_momentum_2020} by freezing the learned representation and adding a linear layer to predict the surface class. We observe that our method outperforms the baselines on all metrics, indicating that our representation can effectively cluster the tactile signals based on the geometric features of the contact surfaces.

\begin{table}[t!]
\centering
\caption{Surface Classification Performance in Real-Robot Experiments.}
\label{tab: surface_class}
\begin{tabular}{c | r | r | r} 
 \hline
 Method & RI $\uparrow$ & ARI $\uparrow$ & Acc $\uparrow$ \\
 \hline
 BYOL & 0.546 & 0.091 & 83.5 \\
 AE & 0.502 & 0.004 & 83.5 \\
 Raw & 0.520 & 0.038 & 72.8 \\
 \textbf{Ours} & \textbf{0.586} & \textbf{0.171} & \textbf{85.4} \\
 \hline
\end{tabular}
\end{table}

\section{Conclusions}

In this paper, we presented a novel framework, HyperTaxel, for learning a geometrically-informed representation of taxel-based tactile signals to achieve hyper-resolution of contact surfaces between the sensor and the object. We introduced a graph-based representation of tactile signals and a contrastive learning objective to learn a correspondence between the low-resolution taxel signals and the high-resolution contact surfaces. We proposed a multi-contact localization algorithm to reduce the uncertainty and ambiguity in the taxel signals and map them to the object surface geometry. We conducted extensive experiments on synthetic and also presented real-world experiments and showed that our framework outperforms the baselines.
We demonstrated that the learned representation can capture the geometric features of the contact surface and generalize across different objects and taxel arrangements. We also showed that the hyper-resolution algorithm can improve the performance of the visuotactile pose estimation model and enable robust sim-to-real transfer.

Our framework opens up new possibilities for leveraging taxel-based tactile sensors for dexterous manipulation and perception. Some future directions for improving the framework include incorporation of temporal information, expanding the contact database, and applying the framework to other modalities.

\footnotesize{
\bibliographystyle{IEEEtranN}
\bibliography{custom, hongyu_zotero}
}

\end{document}